\documentclass[runningheads]{llncs}
\usepackage[T1]{fontenc}
\usepackage{graphicx}
\usepackage{hyperref}
\usepackage{color}

\usepackage[utf8]{inputenc}
\usepackage{mathtools}
\usepackage{multirow}
\usepackage{rotating}
\usepackage{xcolor}
\usepackage{siunitx}
\usepackage[style=lncs,backend=biber,sorting=none,citestyle=numeric-comp]{biblatex}
\usepackage{amsmath,amssymb}
\DeclareMathOperator*{\argmin}{arg\,min}
\usepackage{booktabs}

\DeclareMathOperator{\EX}{\mathbb{E}}

\newtheorem{heuristic}{Heuristic}

\newenvironment{TTP}{%
    \par\noindent\textbf{Target Transformation Problem.} \itshape
}{%
    \par
}

\begin{document}
\title{The When and How of Target Variable Transformations}
%
%
\author{Loren Nuyts\inst{1}\orcidID{0000-0002-4479-3781} \and
Jesse Davis\inst{2,3}\orcidID{0000-0002-3748-9263}}
\authorrunning{L. Nuyts et al.}
%
\institute{KU Leuven, Dept. Computer Science, Research unit DTAI, B-3000 Leuven, Belgium
\email{firstname.lastname@kuleuven.be}}
\maketitle              
\begin{abstract}
The machine learning pipeline typically involves the iterative process of (1) collecting the data, (2) preparing the data, (3) learning a model, and (4) evaluating a model. Practitioners recognize the importance of the data preparation phase in terms of its impact on the ability to learn accurate models. In this regard, significant attention is often paid to manipulating the feature set (e.g., selection, transformations, dimensionality reduction).  A point that is less well appreciated is that transformations on the target variable can also have a large impact on whether it is possible to learn a suitable model. These transformations may include accounting for subject-specific biases (e.g., in how someone uses a rating scale), contexts (e.g., population size effects), and general trends (e.g., inflation). However, this point has received a much more cursory treatment in the existing literature. The goal of this paper is three-fold. First, we aim to highlight the importance of this problem by showing when transforming the target variable has been useful in practice. Second, we will provide a set of generic ``rules of thumb'' that indicate situations when transforming the target variable may be needed. Third, we will discuss which transformations should be considered in a given situation.

\end{abstract}
\section{Introduction}

The machine learning pipeline typically involves the iterative process of (1) collecting the data, (2) preparing the data, (3) learning a model, and (4) evaluating a model. While the machine learning literature primarily focuses on learning algorithms, there is still substantial interest in the steps to prepare or transform the data prior to applying a learning algorithm. Practitioners also recognize the importance of this step in terms of both the time it requires~\cite{preprocessing_scholarly_big_data} and its impact on the quality of the learned model~\cite{big_data_preprocessing, preprocessing_gait_classification}. 

During the data preparation phase, significant attention is often paid to manipulating the feature set. There are three broad categories of operations that are typically considered. First, a variety of techniques exist for transforming an individual feature's values. These approaches range from simple ideas such as standardization or min-max normalization to more complex domain-dependent operations such as removing unwanted biases~\cite{contextuals_gait, context_norm_gas_turbines}. Second, there has been significant work in areas such as feature selection~\cite{feature_selection} and dimensionality reduction~\cite{dimensionality_reduction}. Third, there is work on automatically transforming the given input space to a more optimal representation by applying techniques such as automated feature construction~\cite{davis:viewlearning,tsfuse}, representation learning~\cite{autoencoder_representation_learning, graph_representation_learning}, or manifold learning~\cite{meilua2024manifold, manifold_introduction}. The relevance of the data preparation is also reinforced by the fact that there have been significant efforts made to \emph{automate} this part of the machine learning pipeline: many automated machine learning toolkits explicitly consider including these operations when trying to search for the best configuration of a learning algorithm~\cite{autosklearn, autoencoder_representation_learning}. For example, \textit{Auto-sklearn}~\cite{autosklearn, autosklearn2} automatically decides whether to include scaling, imputation, balancing, or PCA factorization.

Interestingly, \emph{transforming the target variable} can also play a crucial role in facilitating learning a model. As an illustrative example, consider the task of predicting a ship's arrival time at a lock based on historical arrival time data. Using the historical arrival time as the target variable is bad because it strongly depends on the departure time, which in turn is influenced by factors like the crews' shifts and the availability of equipment. Instead, one should \emph{transform the target variable by computing the ship's travel time} and predict that instead~\cite{TTP}.  Similarly, one of the most practically impactful insights from the Netflix challenge was the importance of isolating the user-movie interaction component of a rating by deriving a new variable that controlled for factors related to the user, the movie, and time~\cite{DBLP:conf/kdd/Koren08,koren2010collaborative}. The necessity of deriving a new target variable has also arisen in Kaggle competitions~\cite{wind_masters} and problems like predicting running fatigue~\cite{op2018fatigue}. 

Despite the importance of selecting the appropriate form of the target variable, it has received a much more cursory treatment in existing literature. With a few exceptions (e.g.,~\cite{hyndman2018forecasting}), existing work typically mentions manipulations to the target variable briefly (if at all) without an attempt to understand the general task properties that lead to the need for the transformation~\cite{op2018fatigue, fast_gbdt, DBLP:conf/kdd/Koren08, transformed_target_regressor}. Consequently, we lack a systematic description of the problem characteristics that indicate when a transformation may be useful and what transformations should be considered. The goal of this paper is to fill this gap. Specifically, our contributions are: (1) providing a set of ``rules of thumb'' for when a given target variable might be unsuitable, (2) describing possible mathematical transformations that can be useful in each situation, (3) conducting a set of experiments that demonstrate the utility of the transformation, and (4) providing open-source code for performing them.\footnote{\url{https://github.com/ML-KULeuven/target_transformations}}

\section{Preliminaries and related work}
We assume an $d$-dimensional input space $\mathcal{X} \subset \mathbb{R}^{d}$ and an output space $\mathcal{Y} \subset \mathbb{R}$. 
In this work, we tackle the problem of transforming the target variable. Intuitively, the idea is that applying a function $f$ to observed the target values and training a model to predict $f(\mathcal{Y})$ may be easier than learning a model to directly predict $\mathcal{Y}$. This problem setting can be formalized as follows: 

\begin{TTP}
     Given a dataset $\mathcal{D}_ = \{(\mathbf{x}_1, y_1) \dots (\mathbf{x}_n,y_n)\}$, a model class $\mathcal{H}: \mathcal{X} \mapsto \mathcal{Y}$, and a loss function $\mathcal{L}: \mathcal{Y} \times \mathcal{Y} \mapsto \mathbb{R}$, the target transformation problem aims to find a bijective transformation $f : \mathcal{Y} \to \mathcal{Y}$, s.t. 
     \begin{equation}
     \label{eq:loss}
        \frac{1}{n} \sum_{i=1}^{n} \mathcal{L}(f^{-1}(\hat{h}_{*}(\mathbf{x}_i)), y_{i}) < \frac{1}{n}\sum_{i=1}^{n} \mathcal{L}(\hat{h}(\mathbf{x}_{i}),y_{i})
     \end{equation}
     where we use the inverse transformation because some losses are scale-dependent (e.g., mean absolute error, mean squared error), and $\hat{h}_{*}$ and $\hat{h}$ are trained to predict respectively the transformed target variable and the given target variable:
     \begin{eqnarray*}
          \hat{h}_{*} &=& \argmin_{h \in \mathcal{H}} \frac{1}{n}\sum_{i=1}^{n} \mathcal{L}(h(\mathbf{x}_i)), f(y_{i})) \\
          \hat{h} &=& \argmin_{h \in \mathcal{H}} \frac{1}{n}\sum_{i=1}^{n} \mathcal{L}(h(\mathbf{x}_i)), y_{i}))
     \end{eqnarray*}
\end{TTP}

While prior work~\cite{op2018fatigue, DBLP:conf/kdd/Koren08, fast_gbdt} briefly mentions target transformations, \textcite{hyndman2018forecasting} offer a more detailed discussion of common transformations in forecasting: calendar adjustments, population adjustments, inflation adjustments, and mathematical transformations. Building on this, we (1) synthesize these transformations into three high-level categories, (2) generalize them beyond forecasting, (3) propose additional approaches for the listed dependencies, (4) identify and address additional dependencies that were not previously discussed, and (5) provide a publicly available package for applying these transformations.

\textit{Sklearn} also provides a \textit{TransformedTargetRegressor}\footnote{\url{https://scikit-learn.org/stable/modules/generated/sklearn.compose.TransformedTargetRegressor.html}} that trains a given regression model on a transformed target by applying a user-defined transformation. However, the choice of whether and which transformation to apply is left to the user. Commonly, a logarithmic transformation or a \textit{QuantileTransformer}\footnote{\url{https://scikit-learn.org/stable/modules/generated/sklearn.preprocessing.QuantileTransformer.html}} is used~\cite{transformed_target_regressor}, but, to our knowledge, no study has systematically examined the impact of these transformations on the predictive performance. Notably, this has been neglected by the AutoML literature: \textit{Auto-sklearn}~\cite{autosklearn,autosklearn2} does not support performing target variable transformations and the book on Automated Machine Learning~\cite{hutter2019automated} does not discuss this problem.

\section{A Guide to Target Transformations}
\label{section:transform}
In this section, we will provide motivating examples of when target transformations have been useful in practice. Specifically, we have synthesized the transformations from the literature into three high-level categories and formulated five heuristic ``rules of thumb'' to help practitioners identify characteristics that suggest when a transformation might be beneficial. Finally, within each category, we will suggest several possible transformations.

\subsection{Subjective Target Variable}
Often the target variable is a subjective rating such as how much a user likes a movie (e.g., number of stars given)~\cite{DBLP:conf/kdd/Koren08}, an athlete's rating of perceived exertion (which can be a proxy for fatigue)~\cite{op2018fatigue}, or a person's perceived stress level~\cite{perceived_stress}. Comparing values across subjects can be difficult because they may inherently use the rating scale differently or unmeasured external factors (e.g., mood, baseline fatigue level) can affect a person's rating. 
Therefore, performing subject-level transformations of the target variable can enable learning a more accurate model. 

\begin{heuristic}[Subjective Target]
    The target variable is subjective if it depends on someone's judgment (e.g., rating or questionnaire-derived label) meaning that for subject $i$, $P(y) \neq P(y|i)$. Additionally, the target should be ordinal to allow for transforming it.
\end{heuristic}

   The most straightforward way to normalize a subjective target variable is to center it w.r.t. the mean target of that person (e.g., as was done for Netflix ratings~\cite{DBLP:conf/kdd/Koren08}). This changes the task from predicting the absolute rating to predicting the subject's deviation from their average:
   \begin{equation}
    f(y_{i,j}) = y_{i,j} - \overline{y_i} \quad \text{with} \quad
    \overline{y_i} = \frac{\sum_{j=0}^{S_i} y_{i,j}}{S_i}
\end{equation}
where $y_{i,j}$ is the $j^{th}$ sample of subject $i$ and $S_i$ total number of samples for subject $i$.

The previous transformation accounts for differences between subjects, but there can also be within-subject variation. This can be particularly acute when subjects give a sequence of responses in close succession. For example, Koren~\cite{koren2010collaborative} discusses how a user's movie rating may be given in relation to other recently rated films. Similarly, the evolution of a runner's fatigue during a workout will depend on how tired they are at the start~\cite{op2018fatigue}. In these cases, a local within-subject transformation may be needed. However, determining the appropriate transformation is inherently more challenging because it will rely on the nature of the ratings. If a subject performs several trials where they provide multiple ratings within a trial, it is possible to independently min-max normalize all target variables per trial in order to eliminate any trial-specific effects (e.g., the athlete's starting fatigue state).

\subsection{Contextual dependencies}
Often, there are contextual factors that affect a target's value. To give an illustrative example, consider the production of solar energy. This depends on the number of daylight hours which is determined by two important contextual variables: the season and the latitude of the location. Consequently, production numbers across time and place are not directly comparable. Therefore, it may be simpler to transform the target to solar energy production per daylight hour to remove these effects from the data. We have identified three high-level categories of contexts that may arise: frame of reference, trends, and domain-dependent contexts.

\subsubsection{Frame Dependency} A target variable is often constructed by \textit{aggregating events within a semantically meaningful frame of reference (i.e., context)}. This may be a time frame such as a month or a sports game. It could also be a geographic reference such as a country or region. However, these aggregates are often not directly comparable~\cite{vaep, davis2024methodology, hyndman2018forecasting}.  In the case of time,  months have a varying number of days. Even if the time frame is of a fixed duration, issues can arise: matches within a competition for a sport have a standardized time\footnote{Some obvious caveats exist such as extra time, and soccer's subjective time added.} but the number of points scored depends on the pace (i.e., how quickly teams play)~\cite{davis2024methodology, kubatko:basketball}. Geographic areas vary in terms of their population and size (i.e., land area). An increase in the number of hospital beds for example may reflect population growth rather than a true increase in capacity. To correctly evaluate changes in hospital bed availability, the number of beds per capita should be considered~\cite{hospital_beds, hyndman2018forecasting}. Similarly, crop yield depends on both the yield per plant and the number of plants per unit area~\cite{Bleasdale1960-tu}. 

\begin{heuristic}[Frame Dependency]
    A frame dependency arises if (1) the target variable is aggregated over a frame of reference and (2) its value depends on the choice of this frame of reference.
\end{heuristic}

In these cases, it is natural to normalize the target variable w.r.t. the frame of reference $R$ (e.g., time, area, number of possessions, number of sunlight hours):
\begin{equation}
        f(y_{i}) = \frac{y_{i}}{R}
    \end{equation}
Practically, this transformation normalizes the target to, e.g., sales per day, points per fixed number of possessions in sports~\cite{kubatko:basketball}, or solar energy production per daylight hour.

\subsubsection{Trend Dependency} A common form of historical data involves examples that are collected over an extended period of time. For example, this could be house pricing data, transfers of soccer players, or measurements of stellar noise. In these cases, part of the value of a target variable is explained by a trend that arises due to the passage of time such as inflation~\cite{hyndman2018forecasting}, seasonal fluctuations~\cite{seasonal_trends}, or stellar trends~\cite{stellar_trends}.

\begin{heuristic}[Trend Dependency]
    A trend dependency arises if (1) the target is (partially) explained by a monotonically changing variable, and (2) it is tracked over a sufficiently long time such that the explanatory variable has a significant influence on the target.
\end{heuristic}
What trends may be present is highly domain-dependent. Consequently, picking the best way to transform the target to remove or mitigate the effect of the trend depends on domain knowledge. For monetary values, a common approach is to transform the price to be expressed as a specific year's ($t_0$) cost:
\begin{equation}
        f(y_{i,t}) = \frac{y_{i,t}}{z_t}*z_{{t_0}}
\end{equation}
where $y_{i,t}$ represents the target of subject $i$ at time $t$, and $z_t$ is the price index at time~$t$.

Several trend estimation and detrending techniques have been proposed~\cite{trend_extraction, detrending_methods, detrending_algorithms}. The most appropriate technique often depends on the exact application. However, a detailed discussion is beyond the scope of this paper.

\subsubsection{Specific Domain Knowledge} 
A target variable is often influenced by several domain-specific external factors. For example, a person's stride length is proportional to their height~\cite{contextuals_gait}. Similarly, how much a machine vibrates also depends on the amount of vibrations in nearby machines~\cite{context_norm_gas_turbines}. While some of these causal relationships are known, they are often left unmodeled.
\begin{heuristic}[General Contextual Dependency]
    A general contextual dependency arises if (1) the target depends on external factors that are known in advance, and (2) the relationship between the target and the external factors is not modeled explicitly.
\end{heuristic}
Transforming the target to account for these factors helps eliminate part of its variability. Consequently, it allows the learned model to focus on capturing more complex patterns. We provide two methods to transform the target $y_i$ w.r.t. $k$ contextual variables $\varphi_{i,k}$. The first, proposed by \textcite{context_norm_gas_turbines} for predicting faulty gas turbines, removes the context of a healthy turbine from all dataset features. While originally applied to features, this method can also be extended to the target variable by subtracting its expected value given the contextual variables and dividing by the expected variation $\sigma_i$:
\begin{equation}
    f(y_i) = \frac{y_i - \EX(y_i| \varphi_{i,1} \dotsc \varphi_{i,k})}{\EX(\sigma_i| \varphi_{i,1} \dotsc  \varphi_{i,k})}
\end{equation}
The second method, proposed by \textcite{contextuals_gait}, first fits a multiple regression model to the labels:
\begin{equation}
    y_i = \beta_0 + \sum_{j=1}^{k} \beta_j \varphi_{i,j} + \epsilon_i
\end{equation}
where $\beta_0$ represents the linear regression line intercept term, $\beta_j$ represents the coefficient for the $j^{th}$ contextual feature, and $\epsilon_i \sim N(0,\sigma^2)$ represents the independent and identically distributed residual error for the $i^{th}$ instance. Then, they normalize the gait features by dividing the value of the original feature by the value predicted by the multiple regression model. When applied to a target variable, this becomes:
\begin{equation}
    f(y_i) = \frac{y_i}{\beta_0 + \sum_{j=1}^{k} \beta_j \varphi_{i,j}}
\end{equation}
where the $\beta_j$s are the fitted coefficients from the multiple regression model.

\subsection{Unsuitable Target Distribution}

The distributional characteristics of the target values can also pose problems for learning. On the one hand, models make assumptions about the target variable, such as being homoscedastic, which are often violated in practice~\cite{homoscedasticity_assumption, heteroscedasticity_masterthesis}. On the other hand, regression approaches struggle with imbalance and yield less accurate predictions for rare values or in situations where there are large gaps between target values~\cite{imbalanced_regression1, imbalanced_regression2}.

\begin{heuristic}[Unsuitable Target Distribution]
    An unsuitable target distribution is either heteroscedastic or imbalanced, meaning that specific ranges of target values have significantly fewer observations than others.
\end{heuristic}
To mitigate heteroscedasticity, variance-stabilizing transformations such as the logarithm, square root, or Box-Cox transformations can be applied to normalize the variance~\cite{heteroscedasticity_masterthesis, box1964analysis}.

To address an imbalanced target distribution, one can either (1) incorporate the imbalance into the model or loss function~\cite{imbalanced_regression2}, or (2) transform the target to achieve a more balanced distribution. As the first approach does not involve direct target transformation, it lies outside the scope of this paper. For the second approach, transformations such as the logarithm~\cite{fast_gbdt} are widely used to address skewness. Transforming the target value's distribution to approximate a normal distribution can also mitigate imbalance, as shown in the experiments. 

\begin{table}[tb]
\centering
    \caption{For each dataset, the used acronym, number of instances $n$, the number of features $d$, and the Fisher-Pearson coefficient of skewness $\gamma$ are given. It also indicates the reason for imbalance in the dataset (skewness or gap of missing values).}
    \label{tab:datasets}
\begin{tabular}{l|crrcc}
\toprule
\multicolumn{1}{c|}{\textbf{Dataset full name}} & \textbf{Acronym} & \textbf{$n$} & \textbf{$d$} & \textbf{Skewed/Gap} & \textbf{$\gamma$} \\ \midrule
Auto MPG & AMPG & 398 & 7 & Skewed & 0.455 \\
Bike Sharing & BS & 17389 & 13 & Skewed & 1.277 \\
Combined Cycle Power Plant & CCPP & 9568 & 4 & Gap & 0.306 \\
Concrete Compressive Strength & CCS & 1030 & 8 & Skewed & 0.416 \\
Energy Efficiency: Heating Load & EF1 & 768 & 8 & Gap & 0.360 \\
Energy Efficiency: Cooling Load & EF2 & 768 & 8 & Gap & 0.360 \\
Liver Disorders & LD & 345 & 5 & Skewed & 1.537 \\
Online News Popularity & ONP & 39797 & 58 & Skewed & \hspace{-0.5em}33.963 \\
Real Estate Valuation & REV & 414 & 6 & Skewed & 0.598 \\
Servo & SRV & 167 & 4 & \phantom{l}Skewed + Gap\phantom{l} & 1.775 \\
\bottomrule
\end{tabular}
\end{table}
\section{Experiments}

The previously referenced literature contains numerous examples that empirically demonstrate the benefits of both the subjective~\cite{DBLP:conf/kdd/Koren08,op2018fatigue} and contextual transformations~\cite{kubatko:basketball, davis2024methodology, vaep}. Instead of replicating those experiments, we will focus on transformations related to unsuitable target distributions as to our knowledge there currently is not a systematic empirical study of these. All code and results are available at \url{https://github.com/ML-KULeuven/target_transformations}.

\subsubsection{Setup} We evaluate the performance of a Lasso (\textit{sklearn's Lasso}), Ridge (\textit{sklearn's Ridge}), Gradient Boosted Trees (\textit{sklearn's GradientBoostingRegressor}), and Support Vector (\textit{sklearn's SVR}) regression models on 10 public datasets with an imbalanced distribution. Table~\ref{tab:datasets} shows (1) the general characteristics of the datasets, (2) whether it is skewed, has a gap of missing values, or both, and (3) the Fisher-Pearson coefficient of skewness.

We compare the baseline regression model, without transformations, with the following 4 transformations:
\begin{enumerate}
    \item \textbf{Quantile Normal (QN)} transforms the target with \textit{sklearn's QuantileTransformer} to an approximately normal distribution.
    \item \textbf{Quantile Uniform (QU)} transforms the target with \textit{sklearn's QuantileTransformer} to an approximately uniform distribution.
    \item \textbf{Yeo-Johnson (YJ)} transforms the target with \textit{sklearn's PowerTransformer} to also approximate a normal distribution.
    \item The \textbf{natural logarithm (Ln)} transformation applies $ln(y_i+\text{\emph{offset}})$ to the target, where $\text{\emph{offset}}=\max\limits_{i}(\lceil -y_i \rceil,1)$ ensures that all $y_i > 0$.
\end{enumerate}
Each experiment ran on a single thread on an x86\_64 Intel(R) Xeon(R) CPU @3.20GHz with 32Gb RAM. To mitigate the randomness, we repeat all steps using a five-times repeated two-fold-cross-validation (\textsc{5x2cv}) approach~\cite{5x2cv} and report the average and standard deviation across all folds.

\subsubsection{Evaluation metrics} We employ two evaluation metrics: the relative squared error (RSE) and the symmetric mean absolute percentage error (SMAPE), which are defined as follows:
\begin{equation}
    \text{RSE} = \frac{\sum_{i=1}^{n} (y_i - \hat{y}_i)^2}{\sum_{i=1}^{n} (y_i - \bar{y})^2}
\end{equation}
\begin{equation}
    \text{SMAPE} = \frac{1}{n} \sum_{i=1}^{n} \frac{|y_i - \hat{y}_i|}{\frac{|y_i| + |\hat{y}_i|}{2}} \times 100\%
\end{equation}
where $y_i$ is the actual value, $\bar{y}$ the mean of the actual values, $n$ the number of instances, and $\hat{y}_i$ corresponds to either $\hat{h}(\mathbf{x}_i)$ or $\hat{h}_{*}(\mathbf{x}_i)$ if the model was trained on respectively the original targets or the transformed targets. 

\begin{table}[!ht]
    \centering\caption{Avg. $\pm$ std. RSE ($\downarrow$) across all folds for the Lasso, Ridge, Gradient Boosted Trees (GBTR), and Support Vector Regressor (SVR) regression models, comparing the baseline without any target transformations (Base), and after applying the Quantile Normal (QN), Quantile Uniform (QU), Yeo-Johnson (YJ), and logarithmic (Ln) transformations.}
    \label{tab:rse}
    \begin{tabular}{p{0.5cm}l|l|llll}
\toprule
 &  & \multicolumn{1}{c|}{\textbf{Base}} & \multicolumn{1}{c}{\textbf{QN}} & \multicolumn{1}{c}{\textbf{QU}} & \multicolumn{1}{c}{\textbf{YJ}} & \multicolumn{1}{c}{\textbf{Ln}} \\
\midrule
\multirow{10}{*}{\rotatebox{90}{\textbf{Lasso}}} & AMPG & \num{0.241} {\tiny \textcolor{gray}{$\pm$ \num{0.02}}} & \textbf{0.207} {\tiny \textcolor{gray}{$\pm$ \num{0.03}}} & \num{0.279} {\tiny \textcolor{gray}{$\pm$ \num{0.03}}} & \num{0.219} {\tiny \textcolor{gray}{$\pm$ \num{0.02}}} & \num{0.213} {\tiny \textcolor{gray}{$\pm$ \num{0.02}}} \\
 & BS & \num{1.601} {\tiny \textcolor{gray}{$\pm$ \num{0.07}}} & \num{1.393} {\tiny \textcolor{gray}{$\pm$ \num{0.06}}} & \num{1.91} {\tiny \textcolor{gray}{$\pm$ \num{0.1}}} & \num{1.494} {\tiny \textcolor{gray}{$\pm$ \num{0.05}}} & \textbf{1.238} {\tiny \textcolor{gray}{$\pm$ \num{0.05}}} \\
 & CCPP & \num{0.077} {\tiny \textcolor{gray}{$\pm$ <0.01}} & \textbf{0.073} {\tiny \textcolor{gray}{$\pm$ <0.01}} & \num{0.08} {\tiny \textcolor{gray}{$\pm$ <0.01}} & \num{50.964} {\tiny \textcolor{gray}{$\pm$ \num{52.31}}} & \num{0.075} {\tiny \textcolor{gray}{$\pm$ <0.01}} \\
 & CCS & \num{0.682} {\tiny \textcolor{gray}{$\pm$ \num{0.05}}} & \num{0.625} {\tiny \textcolor{gray}{$\pm$ \num{0.04}}} & \num{0.906} {\tiny \textcolor{gray}{$\pm$ \num{0.07}}} & \num{0.666} {\tiny \textcolor{gray}{$\pm$ \num{0.04}}} & \textbf{0.621} {\tiny \textcolor{gray}{$\pm$ \num{0.04}}} \\
 & EF1 & \num{0.095} {\tiny \textcolor{gray}{$\pm$ \num{0.01}}} & \num{0.088} {\tiny \textcolor{gray}{$\pm$ \num{0.01}}} & \textbf{0.087} {\tiny \textcolor{gray}{$\pm$ \num{0.01}}} & \num{0.093} {\tiny \textcolor{gray}{$\pm$ \num{0.01}}} & \num{0.094} {\tiny \textcolor{gray}{$\pm$ \num{0.01}}} \\
 & EF2 & \num{0.131} {\tiny \textcolor{gray}{$\pm$ \num{0.01}}} & \textbf{0.121} {\tiny \textcolor{gray}{$\pm$ \num{0.01}}} & \num{0.131} {\tiny \textcolor{gray}{$\pm$ \num{0.01}}} & \num{0.129} {\tiny \textcolor{gray}{$\pm$ \num{0.01}}} & \num{0.129} {\tiny \textcolor{gray}{$\pm$ \num{0.01}}} \\
 & LD & \num{17.172} {\tiny \textcolor{gray}{$\pm$ \num{21.18}}} & \textbf{3.935} {\tiny \textcolor{gray}{$\pm$ \num{2.19}}} & \num{7.48} {\tiny \textcolor{gray}{$\pm$ \num{8.7}}} & \num{4.776} {\tiny \textcolor{gray}{$\pm$ \num{2.09}}} & \num{5.303} {\tiny \textcolor{gray}{$\pm$ \num{2.41}}} \\
 & ONP & \num{21.59} {\tiny \textcolor{gray}{$\pm$ \num{21.52}}} & \num{3.618} {\tiny \textcolor{gray}{$\pm$ \num{2.67}}} & \textbf{2.45} {\tiny \textcolor{gray}{$\pm$ \num{1.4}}} & \num{11.554} {\tiny \textcolor{gray}{$\pm$ \num{18.14}}} & \num{20.701} {\tiny \textcolor{gray}{$\pm$ \num{18.54}}} \\
 & REV & \num{0.867} {\tiny \textcolor{gray}{$\pm$ \num{0.26}}} & \textbf{0.781} {\tiny \textcolor{gray}{$\pm$ \num{0.31}}} & \num{0.978} {\tiny \textcolor{gray}{$\pm$ \num{0.31}}} & \num{0.847} {\tiny \textcolor{gray}{$\pm$ \num{0.24}}} & \num{0.823} {\tiny \textcolor{gray}{$\pm$ \num{0.21}}} \\
 & SRV & \num{2.666} {\tiny \textcolor{gray}{$\pm$ \num{3.71}}} & \num{1.666} {\tiny \textcolor{gray}{$\pm$ \num{2.49}}} & \textbf{1.357} {\tiny \textcolor{gray}{$\pm$ \num{0.88}}} & \num{2.3} {\tiny \textcolor{gray}{$\pm$ \num{1.21}}} & \num{2.795} {\tiny \textcolor{gray}{$\pm$ \num{2.43}}} \\
\hline 
\multirow{10}{*}{\rotatebox{90}{\textbf{Ridge}}} & AMPG & \num{0.242} {\tiny \textcolor{gray}{$\pm$ \num{0.03}}} & \textbf{0.197} {\tiny \textcolor{gray}{$\pm$ \num{0.03}}} & \num{0.283} {\tiny \textcolor{gray}{$\pm$ \num{0.04}}} & \num{0.219} {\tiny \textcolor{gray}{$\pm$ \num{0.02}}} & \num{0.213} {\tiny \textcolor{gray}{$\pm$ \num{0.02}}} \\
 & BS & \num{1.582} {\tiny \textcolor{gray}{$\pm$ \num{0.06}}} & \num{1.359} {\tiny \textcolor{gray}{$\pm$ \num{0.04}}} & \num{1.877} {\tiny \textcolor{gray}{$\pm$ \num{0.09}}} & \num{1.471} {\tiny \textcolor{gray}{$\pm$ \num{0.04}}} & \textbf{1.206} {\tiny \textcolor{gray}{$\pm$ \num{0.04}}} \\
 & CCPP & \num{0.077} {\tiny \textcolor{gray}{$\pm$ <0.01}} & \textbf{0.072} {\tiny \textcolor{gray}{$\pm$ <0.01}} & \num{0.08} {\tiny \textcolor{gray}{$\pm$ <0.01}} & \textbf{0.072} {\tiny \textcolor{gray}{$\pm$ <0.01}} & \num{0.075} {\tiny \textcolor{gray}{$\pm$ <0.01}} \\
 & CCS & \num{0.683} {\tiny \textcolor{gray}{$\pm$ \num{0.06}}} & \num{0.623} {\tiny \textcolor{gray}{$\pm$ \num{0.05}}} & \num{0.9} {\tiny \textcolor{gray}{$\pm$ \num{0.09}}} & \num{0.665} {\tiny \textcolor{gray}{$\pm$ \num{0.04}}} & \textbf{0.62} {\tiny \textcolor{gray}{$\pm$ \num{0.03}}} \\
 & EF1 & \num{0.095} {\tiny \textcolor{gray}{$\pm$ \num{0.01}}} & \num{0.091} {\tiny \textcolor{gray}{$\pm$ \num{0.01}}} & \textbf{0.086} {\tiny \textcolor{gray}{$\pm$ \num{0.01}}} & \num{0.09} {\tiny \textcolor{gray}{$\pm$ \num{0.01}}} & \num{0.091} {\tiny \textcolor{gray}{$\pm$ \num{0.01}}} \\
 & EF2 & \num{0.131} {\tiny \textcolor{gray}{$\pm$ \num{0.01}}} & \textbf{0.12} {\tiny \textcolor{gray}{$\pm$ \num{0.01}}} & \num{0.131} {\tiny \textcolor{gray}{$\pm$ \num{0.01}}} & \num{0.127} {\tiny \textcolor{gray}{$\pm$ \num{0.01}}} & \num{0.127} {\tiny \textcolor{gray}{$\pm$ \num{0.01}}} \\
 & LD & \num{3.723} {\tiny \textcolor{gray}{$\pm$ \num{0.75}}} & \textbf{1.827} {\tiny \textcolor{gray}{$\pm$ \num{0.41}}} & \num{3.124} {\tiny \textcolor{gray}{$\pm$ \num{1.22}}} & \num{2.889} {\tiny \textcolor{gray}{$\pm$ \num{0.82}}} & \num{3.119} {\tiny \textcolor{gray}{$\pm$ \num{0.8}}} \\
 & ONP & \num{21.256} {\tiny \textcolor{gray}{$\pm$ \num{21.38}}} & \num{3.503} {\tiny \textcolor{gray}{$\pm$ \num{2.69}}} & \textbf{2.311} {\tiny \textcolor{gray}{$\pm$ \num{1.32}}} & \num{11.573} {\tiny \textcolor{gray}{$\pm$ \num{18.18}}} & \num{21.693} {\tiny \textcolor{gray}{$\pm$ \num{17.22}}} \\
 & REV & \num{0.826} {\tiny \textcolor{gray}{$\pm$ \num{0.24}}} & \textbf{0.721} {\tiny \textcolor{gray}{$\pm$ \num{0.3}}} & \num{0.946} {\tiny \textcolor{gray}{$\pm$ \num{0.32}}} & \num{0.808} {\tiny \textcolor{gray}{$\pm$ \num{0.23}}} & \num{0.766} {\tiny \textcolor{gray}{$\pm$ \num{0.18}}} \\
 & SRV & \num{1.255} {\tiny \textcolor{gray}{$\pm$ \num{0.34}}} & \textbf{0.851} {\tiny \textcolor{gray}{$\pm$ \num{0.42}}} & \num{1.296} {\tiny \textcolor{gray}{$\pm$ \num{0.78}}} & \num{2.246} {\tiny \textcolor{gray}{$\pm$ \num{1.09}}} & \num{2.056} {\tiny \textcolor{gray}{$\pm$ \num{0.72}}} \\
\hline 
\multirow{10}{*}{\rotatebox{90}{\textbf{GBTR}}} & AMPG & \num{0.183} {\tiny \textcolor{gray}{$\pm$ \num{0.03}}} & \num{0.216} {\tiny \textcolor{gray}{$\pm$ \num{0.06}}} & \num{0.201} {\tiny \textcolor{gray}{$\pm$ \num{0.03}}} & \num{0.182} {\tiny \textcolor{gray}{$\pm$ \num{0.03}}} & \textbf{0.181} {\tiny \textcolor{gray}{$\pm$ \num{0.03}}} \\
 & BS & \textbf{0.206} {\tiny \textcolor{gray}{$\pm$ \num{0.01}}} & \num{0.286} {\tiny \textcolor{gray}{$\pm$ \num{0.02}}} & \num{0.342} {\tiny \textcolor{gray}{$\pm$ \num{0.02}}} & \num{0.251} {\tiny \textcolor{gray}{$\pm$ \num{0.02}}} & \num{0.282} {\tiny \textcolor{gray}{$\pm$ \num{0.04}}} \\
 & CCPP & \textbf{0.057} {\tiny \textcolor{gray}{$\pm$ <0.01}} & \num{0.058} {\tiny \textcolor{gray}{$\pm$ <0.01}} & \num{0.062} {\tiny \textcolor{gray}{$\pm$ <0.01}} & \num{50.958} {\tiny \textcolor{gray}{$\pm$ \num{52.31}}} & \textbf{0.057} {\tiny \textcolor{gray}{$\pm$ <0.01}} \\
 & CCS & \num{0.156} {\tiny \textcolor{gray}{$\pm$ \num{0.02}}} & \num{0.153} {\tiny \textcolor{gray}{$\pm$ \num{0.01}}} & \num{0.191} {\tiny \textcolor{gray}{$\pm$ \num{0.02}}} & \textbf{0.152} {\tiny \textcolor{gray}{$\pm$ \num{0.02}}} & \num{0.155} {\tiny \textcolor{gray}{$\pm$ \num{0.02}}} \\
 & EF1 & \textbf{0.003} {\tiny \textcolor{gray}{$\pm$ <0.01}} & \num{0.011} {\tiny \textcolor{gray}{$\pm$ \num{0.01}}} & \num{0.005} {\tiny \textcolor{gray}{$\pm$ <0.01}} & \textbf{0.003} {\tiny \textcolor{gray}{$\pm$ <0.01}} & \textbf{0.003} {\tiny \textcolor{gray}{$\pm$ <0.01}} \\
 & EF2 & \textbf{0.032} {\tiny \textcolor{gray}{$\pm$ <0.01}} & \num{0.041} {\tiny \textcolor{gray}{$\pm$ \num{0.01}}} & \num{0.035} {\tiny \textcolor{gray}{$\pm$ \num{0.01}}} & \textbf{0.032} {\tiny \textcolor{gray}{$\pm$ <0.01}} & \textbf{0.032} {\tiny \textcolor{gray}{$\pm$ <0.01}} \\
 & LD & \num{1.933} {\tiny \textcolor{gray}{$\pm$ \num{0.4}}} & \textbf{1.417} {\tiny \textcolor{gray}{$\pm$ \num{0.18}}} & \num{1.938} {\tiny \textcolor{gray}{$\pm$ \num{0.38}}} & \num{2.056} {\tiny \textcolor{gray}{$\pm$ \num{0.34}}} & \num{2.088} {\tiny \textcolor{gray}{$\pm$ \num{0.33}}} \\
 & ONP & \textbf{11.281} {\tiny \textcolor{gray}{$\pm$ \num{8.56}}} & \num{25.422} {\tiny \textcolor{gray}{$\pm$ \num{8.83}}} & \num{29.567} {\tiny \textcolor{gray}{$\pm$ \num{5.62}}} & \num{42.679} {\tiny \textcolor{gray}{$\pm$ \num{5.72}}} & \num{47.789} {\tiny \textcolor{gray}{$\pm$ \num{6.25}}} \\
 & REV & \num{0.46} {\tiny \textcolor{gray}{$\pm$ \num{0.07}}} & \num{0.442} {\tiny \textcolor{gray}{$\pm$ \num{0.11}}} & \textbf{0.412} {\tiny \textcolor{gray}{$\pm$ \num{0.09}}} & \num{0.463} {\tiny \textcolor{gray}{$\pm$ \num{0.07}}} & \num{0.462} {\tiny \textcolor{gray}{$\pm$ \num{0.08}}} \\
 & SRV & \textbf{0.197} {\tiny \textcolor{gray}{$\pm$ \num{0.1}}} & \num{0.238} {\tiny \textcolor{gray}{$\pm$ \num{0.1}}} & \num{0.226} {\tiny \textcolor{gray}{$\pm$ \num{0.08}}} & \num{0.391} {\tiny \textcolor{gray}{$\pm$ \num{0.27}}} & \num{0.199} {\tiny \textcolor{gray}{$\pm$ \num{0.11}}} \\
\hline 
\multirow{10}{*}{\rotatebox{90}{\textbf{SVR}}} & AMPG & \num{0.393} {\tiny \textcolor{gray}{$\pm$ \num{0.05}}} & \num{0.165} {\tiny \textcolor{gray}{$\pm$ \num{0.02}}} & \num{0.211} {\tiny \textcolor{gray}{$\pm$ \num{0.02}}} & \num{0.165} {\tiny \textcolor{gray}{$\pm$ \num{0.02}}} & \textbf{0.155} {\tiny \textcolor{gray}{$\pm$ \num{0.02}}} \\
 & BS & \num{3.16} {\tiny \textcolor{gray}{$\pm$ \num{0.07}}} & \num{0.826} {\tiny \textcolor{gray}{$\pm$ \num{0.02}}} & \num{0.812} {\tiny \textcolor{gray}{$\pm$ \num{0.04}}} & \num{0.802} {\tiny \textcolor{gray}{$\pm$ \num{0.02}}} & \textbf{0.754} {\tiny \textcolor{gray}{$\pm$ \num{0.02}}} \\
 & CCPP & \num{0.066} {\tiny \textcolor{gray}{$\pm$ <0.01}} & \textbf{0.059} {\tiny \textcolor{gray}{$\pm$ <0.01}} & \num{0.098} {\tiny \textcolor{gray}{$\pm$ \num{0.01}}} & \num{58.833} {\tiny \textcolor{gray}{$\pm$ \num{99.49}}} & \num{40.14} {\tiny \textcolor{gray}{$\pm$ \num{29.91}}} \\
 & CCS & \num{3.428} {\tiny \textcolor{gray}{$\pm$ \num{0.47}}} & \num{0.351} {\tiny \textcolor{gray}{$\pm$ \num{0.03}}} & \num{0.41} {\tiny \textcolor{gray}{$\pm$ \num{0.07}}} & \num{0.367} {\tiny \textcolor{gray}{$\pm$ \num{0.03}}} & \textbf{0.323} {\tiny \textcolor{gray}{$\pm$ \num{0.04}}} \\
 & EF1 & \num{0.126} {\tiny \textcolor{gray}{$\pm$ \num{0.02}}} & \textbf{0.073} {\tiny \textcolor{gray}{$\pm$ \num{0.01}}} & \num{0.097} {\tiny \textcolor{gray}{$\pm$ \num{0.01}}} & \num{0.074} {\tiny \textcolor{gray}{$\pm$ \num{0.01}}} & \num{0.08} {\tiny \textcolor{gray}{$\pm$ \num{0.01}}} \\
 & EF2 & \num{0.177} {\tiny \textcolor{gray}{$\pm$ \num{0.02}}} & \textbf{0.115} {\tiny \textcolor{gray}{$\pm$ \num{0.01}}} & \num{0.125} {\tiny \textcolor{gray}{$\pm$ \num{0.01}}} & \num{0.118} {\tiny \textcolor{gray}{$\pm$ \num{0.01}}} & \num{0.116} {\tiny \textcolor{gray}{$\pm$ \num{0.01}}} \\
 & LD & \num{3.578} {\tiny \textcolor{gray}{$\pm$ \num{1.06}}} & \num{2.0} {\tiny \textcolor{gray}{$\pm$ \num{0.34}}} & \textbf{1.912} {\tiny \textcolor{gray}{$\pm$ \num{0.48}}} & \num{2.233} {\tiny \textcolor{gray}{$\pm$ \num{0.57}}} & \num{2.163} {\tiny \textcolor{gray}{$\pm$ \num{0.52}}} \\
 & ONP & \num{35.717} {\tiny \textcolor{gray}{$\pm$ \num{4.79}}} & \num{8.021} {\tiny \textcolor{gray}{$\pm$ \num{4.59}}} & \textbf{1.874} {\tiny \textcolor{gray}{$\pm$ \num{0.53}}} & \num{37.266} {\tiny \textcolor{gray}{$\pm$ \num{5.01}}} & \num{37.637} {\tiny \textcolor{gray}{$\pm$ \num{5.01}}} \\
 & REV & \num{2.657} {\tiny \textcolor{gray}{$\pm$ \num{0.77}}} & \textbf{0.585} {\tiny \textcolor{gray}{$\pm$ \num{0.09}}} & \num{0.742} {\tiny \textcolor{gray}{$\pm$ \num{0.12}}} & \num{0.689} {\tiny \textcolor{gray}{$\pm$ \num{0.14}}} & \num{0.594} {\tiny \textcolor{gray}{$\pm$ \num{0.12}}} \\
 & SRV & \num{1.931} {\tiny \textcolor{gray}{$\pm$ \num{0.62}}} & \num{0.635} {\tiny \textcolor{gray}{$\pm$ \num{0.45}}} & \num{0.98} {\tiny \textcolor{gray}{$\pm$ \num{0.45}}} & \num{1.378} {\tiny \textcolor{gray}{$\pm$ \num{0.8}}} & \textbf{0.529} {\tiny \textcolor{gray}{$\pm$ \num{0.23}}} \\
\bottomrule
\end{tabular}
\end{table}

\begin{table}[!ht]
\centering
\caption{Avg. $\pm$ std. SMAPE ($\downarrow$) across all folds for the Lasso, Ridge, Gradient Boosted Trees (GBTR), and Support Vector Regressor (SVR) regression models, comparing the baseline without any target transformations (Base), and after applying the Quantile Normal (QN), Quantile Uniform (QU), Yeo-Johnson (YJ), and logarithmic (Ln) transformations.}
    \label{tab:smape}
    \begin{tabular}{p{0.5cm}l|l|llll}
\toprule
 &  & \multicolumn{1}{c|}{\textbf{Base}} & \multicolumn{1}{c}{\textbf{QN}} & \multicolumn{1}{c}{\textbf{QU}} & \multicolumn{1}{c}{\textbf{YJ}} & \multicolumn{1}{c}{\textbf{Ln}} \\
\midrule
\multirow{10}{*}{\rotatebox{90}{\textbf{Lasso}}} & AMPG & \num{11.6} {\tiny \textcolor{gray}{$\pm$ \num{0.48}}} & \num{10.033} {\tiny \textcolor{gray}{$\pm$ \num{0.77}}} & \num{11.062} {\tiny \textcolor{gray}{$\pm$ \num{0.75}}} & \num{9.651} {\tiny \textcolor{gray}{$\pm$ \num{0.46}}} & \textbf{9.44} {\tiny \textcolor{gray}{$\pm$ \num{0.34}}} \\
 & BS & \num{75.372} {\tiny \textcolor{gray}{$\pm$ \num{0.61}}} & \textbf{68.625} {\tiny \textcolor{gray}{$\pm$ \num{0.42}}} & \num{68.8} {\tiny \textcolor{gray}{$\pm$ \num{0.4}}} & \num{68.756} {\tiny \textcolor{gray}{$\pm$ \num{0.3}}} & \num{71.82} {\tiny \textcolor{gray}{$\pm$ \num{0.31}}} \\
 & CCPP & \num{0.8} {\tiny \textcolor{gray}{$\pm$ <0.01}} & \textbf{0.773} {\tiny \textcolor{gray}{$\pm$ <0.01}} & \num{0.819} {\tiny \textcolor{gray}{$\pm$ \num{0.01}}} & \num{2.253} {\tiny \textcolor{gray}{$\pm$ \num{1.21}}} & \num{0.789} {\tiny \textcolor{gray}{$\pm$ <0.01}} \\
 & CCS & \textbf{26.183} {\tiny \textcolor{gray}{$\pm$ \num{0.48}}} & \num{26.401} {\tiny \textcolor{gray}{$\pm$ \num{0.53}}} & \num{27.743} {\tiny \textcolor{gray}{$\pm$ \num{0.6}}} & \num{26.917} {\tiny \textcolor{gray}{$\pm$ \num{0.46}}} & \num{28.641} {\tiny \textcolor{gray}{$\pm$ \num{0.56}}} \\
 & EF1 & \num{9.989} {\tiny \textcolor{gray}{$\pm$ \num{0.31}}} & \num{9.1} {\tiny \textcolor{gray}{$\pm$ \num{0.79}}} & \num{8.627} {\tiny \textcolor{gray}{$\pm$ \num{0.13}}} & \num{8.546} {\tiny \textcolor{gray}{$\pm$ \num{0.27}}} & \textbf{8.533} {\tiny \textcolor{gray}{$\pm$ \num{0.27}}} \\
 & EF2 & \num{9.073} {\tiny \textcolor{gray}{$\pm$ \num{0.31}}} & \num{8.682} {\tiny \textcolor{gray}{$\pm$ \num{0.33}}} & \num{8.599} {\tiny \textcolor{gray}{$\pm$ \num{0.35}}} & \num{8.409} {\tiny \textcolor{gray}{$\pm$ \num{0.21}}} & \textbf{8.401} {\tiny \textcolor{gray}{$\pm$ \num{0.23}}} \\
 & LD & \textbf{83.908} {\tiny \textcolor{gray}{$\pm$ \num{3.92}}} & \num{84.781} {\tiny \textcolor{gray}{$\pm$ \num{2.78}}} & \num{85.488} {\tiny \textcolor{gray}{$\pm$ \num{3.21}}} & \num{87.843} {\tiny \textcolor{gray}{$\pm$ \num{3.25}}} & \num{87.034} {\tiny \textcolor{gray}{$\pm$ \num{3.68}}} \\
 & ONP & \num{77.221} {\tiny \textcolor{gray}{$\pm$ \num{1.1}}} & \textbf{54.842} {\tiny \textcolor{gray}{$\pm$ \num{0.14}}} & \num{55.166} {\tiny \textcolor{gray}{$\pm$ \num{0.14}}} & \num{55.699} {\tiny \textcolor{gray}{$\pm$ \num{0.1}}} & \num{57.154} {\tiny \textcolor{gray}{$\pm$ \num{0.15}}} \\
 & REV & \num{19.372} {\tiny \textcolor{gray}{$\pm$ \num{2.65}}} & \num{17.634} {\tiny \textcolor{gray}{$\pm$ \num{1.93}}} & \num{19.886} {\tiny \textcolor{gray}{$\pm$ \num{2.36}}} & \num{18.162} {\tiny \textcolor{gray}{$\pm$ \num{3.73}}} & \textbf{15.871} {\tiny \textcolor{gray}{$\pm$ \num{0.58}}} \\
 & SRV & \num{77.731} {\tiny \textcolor{gray}{$\pm$ \num{7.02}}} & \num{44.934} {\tiny \textcolor{gray}{$\pm$ \num{4.16}}} & \textbf{41.114} {\tiny \textcolor{gray}{$\pm$ \num{2.89}}} & \num{44.545} {\tiny \textcolor{gray}{$\pm$ \num{2.42}}} & \num{59.915} {\tiny \textcolor{gray}{$\pm$ \num{5.2}}} \\
\hline 
\multirow{10}{*}{\rotatebox{90}{\textbf{Ridge}}} & AMPG & \num{11.691} {\tiny \textcolor{gray}{$\pm$ \num{0.41}}} & \num{10.164} {\tiny \textcolor{gray}{$\pm$ \num{0.81}}} & \num{11.109} {\tiny \textcolor{gray}{$\pm$ \num{0.73}}} & \num{9.668} {\tiny \textcolor{gray}{$\pm$ \num{0.51}}} & \textbf{9.454} {\tiny \textcolor{gray}{$\pm$ \num{0.38}}} \\
 & BS & \num{75.488} {\tiny \textcolor{gray}{$\pm$ \num{0.52}}} & \textbf{68.674} {\tiny \textcolor{gray}{$\pm$ \num{0.41}}} & \num{68.783} {\tiny \textcolor{gray}{$\pm$ \num{0.4}}} & \num{68.732} {\tiny \textcolor{gray}{$\pm$ \num{0.29}}} & \num{71.773} {\tiny \textcolor{gray}{$\pm$ \num{0.32}}} \\
 & CCPP & \num{0.8} {\tiny \textcolor{gray}{$\pm$ <0.01}} & \num{0.775} {\tiny \textcolor{gray}{$\pm$ <0.01}} & \num{0.821} {\tiny \textcolor{gray}{$\pm$ \num{0.01}}} & \textbf{0.773} {\tiny \textcolor{gray}{$\pm$ <0.01}} & \num{0.79} {\tiny \textcolor{gray}{$\pm$ <0.01}} \\
 & CCS & \textbf{26.22} {\tiny \textcolor{gray}{$\pm$ \num{0.47}}} & \num{26.41} {\tiny \textcolor{gray}{$\pm$ \num{0.63}}} & \num{27.743} {\tiny \textcolor{gray}{$\pm$ \num{0.65}}} & \num{26.942} {\tiny \textcolor{gray}{$\pm$ \num{0.48}}} & \num{28.678} {\tiny \textcolor{gray}{$\pm$ \num{0.56}}} \\
 & EF1 & \num{9.988} {\tiny \textcolor{gray}{$\pm$ \num{0.31}}} & \num{9.798} {\tiny \textcolor{gray}{$\pm$ \num{0.39}}} & \textbf{8.626} {\tiny \textcolor{gray}{$\pm$ \num{0.12}}} & \num{8.867} {\tiny \textcolor{gray}{$\pm$ \num{0.27}}} & \num{8.94} {\tiny \textcolor{gray}{$\pm$ \num{0.28}}} \\
 & EF2 & \num{9.071} {\tiny \textcolor{gray}{$\pm$ \num{0.3}}} & \num{8.768} {\tiny \textcolor{gray}{$\pm$ \num{0.25}}} & \num{8.584} {\tiny \textcolor{gray}{$\pm$ \num{0.34}}} & \textbf{8.4} {\tiny \textcolor{gray}{$\pm$ \num{0.22}}} & \num{8.401} {\tiny \textcolor{gray}{$\pm$ \num{0.22}}} \\
 & LD & \num{83.031} {\tiny \textcolor{gray}{$\pm$ \num{3.33}}} & \textbf{82.812} {\tiny \textcolor{gray}{$\pm$ \num{3.12}}} & \num{84.483} {\tiny \textcolor{gray}{$\pm$ \num{2.32}}} & \num{86.545} {\tiny \textcolor{gray}{$\pm$ \num{2.48}}} & \num{85.675} {\tiny \textcolor{gray}{$\pm$ \num{2.78}}} \\
 & ONP & \num{77.306} {\tiny \textcolor{gray}{$\pm$ \num{1.15}}} & \textbf{54.841} {\tiny \textcolor{gray}{$\pm$ \num{0.14}}} & \num{55.161} {\tiny \textcolor{gray}{$\pm$ \num{0.14}}} & \num{55.703} {\tiny \textcolor{gray}{$\pm$ \num{0.1}}} & \num{57.146} {\tiny \textcolor{gray}{$\pm$ \num{0.15}}} \\
 & REV & \num{19.359} {\tiny \textcolor{gray}{$\pm$ \num{2.6}}} & \num{17.619} {\tiny \textcolor{gray}{$\pm$ \num{1.92}}} & \num{19.79} {\tiny \textcolor{gray}{$\pm$ \num{2.37}}} & \num{18.262} {\tiny \textcolor{gray}{$\pm$ \num{3.7}}} & \textbf{15.928} {\tiny \textcolor{gray}{$\pm$ \num{0.67}}} \\
 & SRV & \num{78.886} {\tiny \textcolor{gray}{$\pm$ \num{5.49}}} & \num{44.006} {\tiny \textcolor{gray}{$\pm$ \num{4.02}}} & \textbf{40.979} {\tiny \textcolor{gray}{$\pm$ \num{2.72}}} & \num{44.42} {\tiny \textcolor{gray}{$\pm$ \num{2.35}}} & \num{59.591} {\tiny \textcolor{gray}{$\pm$ \num{4.87}}} \\
\hline 
\multirow{10}{*}{\rotatebox{90}{\textbf{GBTR}}} & AMPG & \num{9.223} {\tiny \textcolor{gray}{$\pm$ \num{0.43}}} & \num{10.06} {\tiny \textcolor{gray}{$\pm$ \num{0.78}}} & \num{9.361} {\tiny \textcolor{gray}{$\pm$ \num{0.44}}} & \num{9.209} {\tiny \textcolor{gray}{$\pm$ \num{0.41}}} & \textbf{9.191} {\tiny \textcolor{gray}{$\pm$ \num{0.45}}} \\
 & BS & \num{45.62} {\tiny \textcolor{gray}{$\pm$ \num{0.77}}} & \num{32.653} {\tiny \textcolor{gray}{$\pm$ \num{0.91}}} & \num{34.414} {\tiny \textcolor{gray}{$\pm$ \num{0.61}}} & \num{31.519} {\tiny \textcolor{gray}{$\pm$ \num{0.97}}} & \textbf{31.347} {\tiny \textcolor{gray}{$\pm$ \num{1.24}}} \\
 & CCPP & \textbf{0.66} {\tiny \textcolor{gray}{$\pm$ \num{0.01}}} & \num{0.671} {\tiny \textcolor{gray}{$\pm$ <0.01}} & \num{0.677} {\tiny \textcolor{gray}{$\pm$ \num{0.01}}} & \num{2.209} {\tiny \textcolor{gray}{$\pm$ \num{1.26}}} & \num{0.661} {\tiny \textcolor{gray}{$\pm$ \num{0.01}}} \\
 & CCS & \num{13.545} {\tiny \textcolor{gray}{$\pm$ \num{0.74}}} & \num{13.757} {\tiny \textcolor{gray}{$\pm$ \num{0.82}}} & \num{15.519} {\tiny \textcolor{gray}{$\pm$ \num{1.05}}} & \textbf{13.155} {\tiny \textcolor{gray}{$\pm$ \num{0.52}}} & \num{13.353} {\tiny \textcolor{gray}{$\pm$ \num{0.67}}} \\
 & EF1 & \num{1.805} {\tiny \textcolor{gray}{$\pm$ \num{0.11}}} & \num{2.992} {\tiny \textcolor{gray}{$\pm$ \num{0.45}}} & \num{2.319} {\tiny \textcolor{gray}{$\pm$ \num{0.33}}} & \num{1.776} {\tiny \textcolor{gray}{$\pm$ \num{0.11}}} & \textbf{1.77} {\tiny \textcolor{gray}{$\pm$ \num{0.13}}} \\
 & EF2 & \num{4.007} {\tiny \textcolor{gray}{$\pm$ \num{0.24}}} & \num{4.537} {\tiny \textcolor{gray}{$\pm$ \num{0.41}}} & \num{4.046} {\tiny \textcolor{gray}{$\pm$ \num{0.26}}} & \textbf{3.803} {\tiny \textcolor{gray}{$\pm$ \num{0.22}}} & \num{3.815} {\tiny \textcolor{gray}{$\pm$ \num{0.21}}} \\
 & LD & \num{87.184} {\tiny \textcolor{gray}{$\pm$ \num{3.71}}} & \num{90.263} {\tiny \textcolor{gray}{$\pm$ \num{3.0}}} & \textbf{85.091} {\tiny \textcolor{gray}{$\pm$ \num{3.75}}} & \num{86.575} {\tiny \textcolor{gray}{$\pm$ \num{3.41}}} & \num{86.818} {\tiny \textcolor{gray}{$\pm$ \num{3.43}}} \\
 & ONP & \num{74.417} {\tiny \textcolor{gray}{$\pm$ \num{0.61}}} & \textbf{53.787} {\tiny \textcolor{gray}{$\pm$ \num{0.13}}} & \num{54.131} {\tiny \textcolor{gray}{$\pm$ \num{0.15}}} & \num{54.809} {\tiny \textcolor{gray}{$\pm$ \num{0.12}}} & \num{56.213} {\tiny \textcolor{gray}{$\pm$ \num{0.13}}} \\
 & REV & \num{14.54} {\tiny \textcolor{gray}{$\pm$ \num{0.95}}} & \num{15.175} {\tiny \textcolor{gray}{$\pm$ \num{1.15}}} & \textbf{13.86} {\tiny \textcolor{gray}{$\pm$ \num{0.63}}} & \num{14.368} {\tiny \textcolor{gray}{$\pm$ \num{0.94}}} & \num{14.475} {\tiny \textcolor{gray}{$\pm$ \num{1.07}}} \\
 & SRV & \num{24.719} {\tiny \textcolor{gray}{$\pm$ \num{3.63}}} & \num{27.944} {\tiny \textcolor{gray}{$\pm$ \num{4.97}}} & \num{24.467} {\tiny \textcolor{gray}{$\pm$ \num{2.63}}} & \num{24.077} {\tiny \textcolor{gray}{$\pm$ \num{1.32}}} & \textbf{22.017} {\tiny \textcolor{gray}{$\pm$ \num{2.0}}} \\
\hline 
\multirow{10}{*}{\rotatebox{90}{\textbf{SVR}}} & AMPG & \num{10.605} {\tiny \textcolor{gray}{$\pm$ \num{0.4}}} & \textbf{8.507} {\tiny \textcolor{gray}{$\pm$ \num{0.52}}} & \num{9.606} {\tiny \textcolor{gray}{$\pm$ \num{0.45}}} & \num{8.511} {\tiny \textcolor{gray}{$\pm$ \num{0.52}}} & \num{8.736} {\tiny \textcolor{gray}{$\pm$ \num{0.45}}} \\
 & BS & \num{68.705} {\tiny \textcolor{gray}{$\pm$ \num{0.46}}} & \num{54.697} {\tiny \textcolor{gray}{$\pm$ \num{0.24}}} & \num{55.962} {\tiny \textcolor{gray}{$\pm$ \num{0.45}}} & \num{55.428} {\tiny \textcolor{gray}{$\pm$ \num{0.26}}} & \textbf{54.447} {\tiny \textcolor{gray}{$\pm$ \num{0.19}}} \\
 & CCPP & \num{0.716} {\tiny \textcolor{gray}{$\pm$ \num{0.01}}} & \textbf{0.686} {\tiny \textcolor{gray}{$\pm$ \num{0.01}}} & \num{0.807} {\tiny \textcolor{gray}{$\pm$ \num{0.02}}} & \num{2.221} {\tiny \textcolor{gray}{$\pm$ \num{1.26}}} & \num{3.356} {\tiny \textcolor{gray}{$\pm$ \num{0.04}}} \\
 & CCS & \num{30.865} {\tiny \textcolor{gray}{$\pm$ \num{0.46}}} & \textbf{19.823} {\tiny \textcolor{gray}{$\pm$ \num{0.53}}} & \num{20.602} {\tiny \textcolor{gray}{$\pm$ \num{0.58}}} & \num{20.389} {\tiny \textcolor{gray}{$\pm$ \num{0.58}}} & \num{20.323} {\tiny \textcolor{gray}{$\pm$ \num{0.76}}} \\
 & EF1 & \num{9.542} {\tiny \textcolor{gray}{$\pm$ \num{0.4}}} & \textbf{7.912} {\tiny \textcolor{gray}{$\pm$ \num{0.31}}} & \num{10.406} {\tiny \textcolor{gray}{$\pm$ \num{0.41}}} & \num{8.105} {\tiny \textcolor{gray}{$\pm$ \num{0.28}}} & \num{9.047} {\tiny \textcolor{gray}{$\pm$ \num{0.2}}} \\
 & EF2 & \num{9.326} {\tiny \textcolor{gray}{$\pm$ \num{0.32}}} & \textbf{7.882} {\tiny \textcolor{gray}{$\pm$ \num{0.39}}} & \num{8.925} {\tiny \textcolor{gray}{$\pm$ \num{0.23}}} & \num{8.031} {\tiny \textcolor{gray}{$\pm$ \num{0.39}}} & \num{9.075} {\tiny \textcolor{gray}{$\pm$ \num{0.32}}} \\
 & LD & \num{84.227} {\tiny \textcolor{gray}{$\pm$ \num{5.28}}} & \textbf{82.418} {\tiny \textcolor{gray}{$\pm$ \num{2.79}}} & \num{82.923} {\tiny \textcolor{gray}{$\pm$ \num{2.43}}} & \num{83.98} {\tiny \textcolor{gray}{$\pm$ \num{3.43}}} & \num{84.141} {\tiny \textcolor{gray}{$\pm$ \num{3.72}}} \\
 & ONP & \num{57.568} {\tiny \textcolor{gray}{$\pm$ \num{0.17}}} & \num{54.582} {\tiny \textcolor{gray}{$\pm$ \num{0.17}}} & \num{54.747} {\tiny \textcolor{gray}{$\pm$ \num{0.16}}} & \textbf{54.528} {\tiny \textcolor{gray}{$\pm$ \num{0.18}}} & \num{54.546} {\tiny \textcolor{gray}{$\pm$ \num{0.18}}} \\
 & REV & \num{19.748} {\tiny \textcolor{gray}{$\pm$ \num{1.82}}} & \num{15.07} {\tiny \textcolor{gray}{$\pm$ \num{0.98}}} & \num{16.434} {\tiny \textcolor{gray}{$\pm$ \num{1.75}}} & \num{15.245} {\tiny \textcolor{gray}{$\pm$ \num{1.0}}} & \textbf{14.638} {\tiny \textcolor{gray}{$\pm$ \num{0.82}}} \\
 & SRV & \num{41.102} {\tiny \textcolor{gray}{$\pm$ \num{3.45}}} & \textbf{31.626} {\tiny \textcolor{gray}{$\pm$ \num{4.04}}} & \num{32.313} {\tiny \textcolor{gray}{$\pm$ \num{3.27}}} & \num{34.047} {\tiny \textcolor{gray}{$\pm$ \num{3.18}}} & \num{34.16} {\tiny \textcolor{gray}{$\pm$ \num{2.68}}} \\
\bottomrule
\end{tabular}
\end{table}

\subsubsection{Results} 
Tables~\ref{tab:rse}~and~\ref{tab:smape} provide the RSE and SMAPE results for all four included regression models. The QN transformation significantly improves performance for Lasso, Ridge, and SVR, yielding median improvements of between 11\% and 63\% for RSE and between 10\% and 19\% for SMAPE. In contrast, the GBTR seems to be adversely affected by the normal distribution with a median increase in RSE of 19\%, and SMAPE of 4\%.

Both the YJ and Ln transformations substantially improve over the baseline for Lasso, Ridge, and SVR, with median improvements for YJ (Ln) in RSE ranging from 5\% to 40\% (5\% to 50\%), and in SMAPE from 8\% to 16\% (9\% to 11\%). The advantages for the GBTR model are less clear: while the median RSE increases with 6\% for Yeo-Johnson and 1\% for the logarithm, the median SMAPE decreases with 2\% for both transformations. 

In contrast to the previous transformations, the QU does not convincingly outperform the baseline for Lasso, Ridge, and GBTR.  Although the median SMAPE improves by between 0\% and 5\%, the median RSE drops by between -2\% and -12\%, which makes it unclear whether these regression models benefit from the uniform distribution. Only SVR consistently benefits from the QU transformation, with a median decrease in RSE of 48\% and in SMAPE of 7\%.

To conclude, applying the QN, YJ, and Ln transformations to an imbalanced target distribution significantly improves the predictions for Lasso, Ridge, and SVR. In contrast, GBTR shows a general robustness to the distribution of the target variable. The linearity of the model likely plays a role in how susceptible the model is to an imbalanced target distribution. Lasso and Ridge are both linear regression models and SVR fits a linear model after applying a kernel function. GBTR on the other hand does not require the target to be linear in any hyperplane of the data, making it more suitable for an imbalanced target distribution.
Finally, the QU transformer only shows considerable improvements for the SVR model and mixed results for the other approaches, indicating that a normal distribution is preferred over a uniform one for these algorithms.

\section{Conclusion}
This paper highlights that transforming the target variable can also be crucial for obtaining accurate models. It provides a systematic description of the characteristics that indicate when a transformation may be beneficial and what transformations should be considered. Specifically, we (1) propose a set of heuristic ``rules of thumb'' that indicate when a target might be unsuitable, (2) describe several mathematical transformations for each situation, (3) empirically evaluate the effect of transforming the target variable's distribution on the predictive performance of four well-known regression approaches, and (4) offer an open-source repository to facilitate applying these transformations. Together, these contributions aim to emphasize the often-overlooked dependencies within target variables and lower the barrier to use target transformations.

\begin{credits}
\subsubsection{\ackname} This work received funding from the Interuniversity Special Research Fund (IBOF/21/075) and the Flemish Government under the “Onderzoeksprogramma Artifciële Intelligentie (AI) Vlaanderen” program. We thank Wannes Meert and Lorenzo Perini for their feedback. 

\subsubsection{\discintname}
The authors have no competing interests to declare that are relevant to the content of this article.
\end{credits}

\printbibliography
%
%
%
%





\end{document}